%% file: paper.tex
\documentclass{article}
\usepackage{spconf,amsmath,graphicx}
\usepackage{amssymb, bm, graphicx, hyperref, enumerate, psfrag, rotating}
\usepackage{verbatim}
\usepackage{subfigure}
\usepackage[boxruled,norelsize]{algorithm2e}
\usepackage{url}

\input{macro+variables}

\makeatletter
\newcommand{\nosemic}{\renewcommand{\@endalgocfline}{\relax}}
\newcommand{\dosemic}{\renewcommand{\@endalgocfline}{\algocf@endline}}
\makeatother

\makeatletter
\newcommand{\removelatexerror}{\let\@latex@error\@gobble}
\makeatother

\title{Compressive k-means}
\newcommand{\INRIA}{{INRIA Rennes - Bretagne Atlantique, Campus de Beaulieu, FR-35042 Rennes Cedex, France}}

\newcommand{\GIPSA}{{CNRS, GIPSA-lab, FR-38402 Saint Martin d'Heres Cedex, France}}
\newcommand{\ERC}{{This work was partly funded by the European Research Council, PLEASE project
(ERC-StG-2011-277906).}}
\newcommand{\UR}{{Universit\'e Rennes 1, FR-35065 Rennes Cedex, France}}
\name{Nicolas Keriven$^{\dagger \star}$ \qquad Nicolas Tremblay$^{\ddagger}$ \qquad Yann Traonmilin$^{\star}$ \qquad R\'emi Gribonval $^{\star}$ \thanks{\ERC}}
\address{$^{\star}$ \INRIA \\
$^{\dagger}$ \UR \\
$^\ddagger$ \GIPSA}
\newcommand{\sca}{0.23}
\newcommand{\vsp}{-10pt}

\begin{document}
\ninept
\maketitle
\begin{abstract}
The Lloyd-Max algorithm is a classical approach to perform $K$-means clustering.
Unfortunately, its cost becomes prohibitive as the training dataset grows large. 
We propose a compressive version of $K$-means (CKM), that estimates cluster centers from a \emph{sketch}, i.e. from a drastically compressed representation of the training dataset.  

We demonstrate empirically that CKM performs similarly to Lloyd-Max,
for a sketch size proportional to the number of centroids times the ambient dimension, and {\em independent of the size of the original dataset}.  
Given the sketch, the computational complexity of CKM is also independent of the size of the dataset.
Unlike Lloyd-Max which requires several replicates, we further demonstrate that CKM is \emph{almost insensitive to initialization}. 
For a large dataset of $10^7$ data points, we show that CKM can run two orders of magnitude faster than five replicates of Lloyd-Max, with similar clustering performance on artificial data. 
Finally, CKM achieves lower classification errors on handwritten digits classification.
\end{abstract}

\section{Introduction}\label{sec:intro}
Given a set of datapoints $\dataset=\{\vx_1,\vx_2,\ldots,\vx_N\}\subset\mathbb{R}^n$ (the dataset), the 
sum of squared errors (SSE) problem is to find 
$K$ centroids $\support=\{\vc_1,\ldots,\vc_K\}\subset \mathbb{R}^n$ such that the SSE is minimized:
\begin{equation}\label{eq:SSE}
 \mbox{SSE}(\dataset,\support)=\sum_{i=1}^N \min_{k} \norm{\vx_i-\vc_k}^2.
\end{equation}
Finding the global minimum of this cost function is NP-hard~\cite{Aloise2009}, and Lloyd~\cite{Lloyd1982} and Steinhaus~\cite{Steinhaus1956} proposed 50 years ago a classical heuristics that is still commonly used today. Its complexity is $O(nNKI)$ with $I$ 
 the number of iterations until convergence. This 
becomes prohibitive when any of these factors become too large. 

In this paper, we propose a heuristics to find the centroids $\support$ from a \emph{sketch} of the dataset $\dataset$ which size $m$ \emph{does not depend} on $N$. More precisely, a sketching procedure $\sk$ that converts a set of weighted vectors in $\Rbb^n$ to a vector in $\Cbb^m$ is defined, and centroids are derived by finding a set $\support$ of weighted points that has a sketch close to that of the dataset $\dataset$ with uniform weights:
\begin{equation}
\label{eq:pbm}
\argmin_{\support,\valpha} \norm{\sk(\dataset,\vone/N) - \sk(C,\valpha)}_2^2
\end{equation}
with $\valpha\geq 0$, $\sum_{k=1}^K \alpha_k=1$. 
As we will see, computing the sketch $\sk(\dataset,\vone/N) \in \Cbb^m$ of the dataset requires only one pass over $\dataset$. 
It can also benefit from distributed or online computing. 


Centroids can be retrieved from the sketch with a variant of Compressive Learning Orthogonal Matching Pursuit (CLOMPR) \cite{Keriven2015,Keriven2016}, an algorithm initially used for large-scale Gaussian Mixture Model (GMM) estimation. Its complexity reads $\order{nmK^2}$, thus eliminating entirely the dependence in $N$ once the sketch has been computed. This complexity could be further reduced by leveraging fast transforms \cite{Do2011,Lemagoarou2015} or embeddings in lower dimension as a preprocessing step \cite{Boutsidis2010}.

We discuss related works in Section \ref{sec:related_work}, and describe the proposed method in Section \ref{sec:algorithm}. Experiments on artificial and real data are performed in Section \ref{sec:experiments}. Even though the cost function \eqref{eq:pbm} bears no immediate connection with the SSE cost~\eqref{eq:SSE}, we show empirically that the SSE obtained with the proposed method approaches that obtained with repeated runs of Lloyd-Max. 
Moreover the proposed algorithm is 
more stable to initialization, and although the sketch size 
is independent of $N$, the proposed algorithm performs much better than repeated runs of Lloyd-Max on large datasets, both in terms of clustering quality and computational complexity, with observed runtimes (\emph{given the sketch}) two orders of magnitude faster for a large dataset of $10^7$ data points.

%

\section{Related Work}\label{sec:related_work}

Several lines of work tackle $K$-means on large datasets. The clever initialization of $K$-means++~\cite{Arthur2007} increases stability of the Lloyd-Max algorithm and decreases the number of iterations $I$ until convergence. 
Some works 
reduce the ambient dimension $n$, either by selecting a limited number of features~\cite{Altschuler2016,Boutsidis2009}, or by embedding all points in a lower dimension using, for instance, random projections~\cite{Boutsidis2010}.

Closer to our work, coresets methods~\cite{Frahling2005, Feldman2011}
~aim at reducing the number of datapoints $N$ by constructing intermediate structures that retain some properties of the SSE. Like the proposed sketches, coresets can also be constructed in a distributed/online fashion \cite{Feldman2011}. 
 Unlike coreset methods our approach does not explicitly aim at approximating the SSE but uses a different objective function.

Our sketching structure bears connection with Random Fourier Features \cite{Rahimi2007} in the context of Hilbert space embedding of probability distributions \cite{Sriperumbudur2010} (see \cite{Keriven2016} for further details). Similar embeddings have been used, for instance, in the context of classification of distributions \cite{Sutherland2015}. Here we will see that the proposed approach can be formulated as an infinite-dimensional Compressive Sensing problem, in which a \emph{probability distribution} is measured through a random linear operator, then decoded under the form of a ``sparse'' distribution, i.e. a finite combination of Diracs whose locations correspond to the desired centroids. This problem can be linked with the super-resolution problem where one aims at estimating combinations of Diracs from a low pass observation. In this case, in dimension one, stable recovery in the low noise regime is possible 
based on the minimization of the total variation of probability measures  \cite{Candes2012}. However, the extension of these techniques to higher dimensions $n\gg1$ does not yield practical results yet~\cite{Castro2015}. The CLOMPR heuristics 
empirically overcomes these limitations.  

\section{Proposed method}\label{sec:algorithm}


In many important applications, one wishes to reconstruct a signal from incomplete samples of its discrete Fourier transform \cite{Candes2006,Candes2006b}. A classical result from Compressive Sensing states that a few randomly selected Fourier samples contain ``enough" information to reconstruct a sparse signal with high probability.

\subsection{Sketching operator}

Given $m$ frequency vectors $\freqs=\{\freq_1,...,\freq_m\}$ in $\Rbb^n$, the sketch of a set of $L$ points $Y$ with weights $\vec{\beta}$ is formed as follows:
\begin{equation}
\label{eq:sketch}
\sk(Y,\vec{\beta})=\Big[\sum_{l=1}^L \beta_l e^{-i\freq_j^T\vy_l}\Big]_{j=1}^m \in \Cbb^m
\end{equation}
This sketching procedure can be reformulated as an operator $\skop$ which is linear \emph{with respect to probability distributions}. Define this operator as a sampling of the characteristic function $\skop p=\left[\Ebb_{\vx\sim p}e^{-i\freq_j^T\vx}\right]_{j=1}^m$ of a probability distribution $p$ at frequencies $\freq_1,...,\freq_m$. Denoting $p_{Y,\vec{\beta}}=\sum_{l=1}^L \beta_l \dirac_{\vy_l}$, the problem \eqref{eq:pbm} can be reformulated as 
\begin{equation}
\label{eq:pbm_skop}
\argmin_{\support,\valpha} \norm{\hat \vz - \skop p_{\support,\valpha}}_2^2
\end{equation}
where $\hat\vz=\skop\hat p_\dataset$ is the sketch of the dataset, with $\hat p_\dataset=\frac{1}{N}\sum_{i=1}^N \dirac_{\vx_i}$ the empirical distribution of the data.

In the spirit of Random Fourier Sampling, the frequencies $\freq_j$ are drawn $i.i.d.$ from a distribution $\freqdist$. In previous work \cite{Keriven2015, Keriven2016}, we proposed a distribution $\freqdist$ referred to as \emph{Adapted radius} frequency distribution, based on a heuristics that maximizes the variation of the characteristic function at the selected frequencies when data are drawn from a GMM. In this paper we show empirically that this distribution is also adapted to a variety of scenarios even outside the context of GMMs. The Adapted radius distribution is parametrized by a scaling quantity $\sigma^2$. In \cite{Keriven2016} an algorithm to choose this scale parameter is proposed, by computing a small sketch of (a fraction of) the dataset $\dataset$ and performing an adapted regression on it.

 The approach thus corresponds to a generalized Compressive Sensing problem, where we measure the probability distribution $\hat p_\dataset$, which is the ``signal" of interest, through a random linear operator $\skop$, and reconstruct it under the form of a ``sparse" distribution $p_{\support,\valpha}$ only supported on a few Diracs.

\subsection{CLOMPR algorithm}

The CLOMPR algorithm 
is a heuristic algorithm to seek a solution to problem \eqref{eq:pbm_skop} that has been 
proposed in previous work \cite{Keriven2015} for Gaussian Mixture Model estimation. It is a greedy algorithm inspired by Orthogonal Matching Pursuit (OMP) and its variant OMP with Replacement (OMPR), which comprises more iterations than OMP, with an additional Hard Thresholding step. As we recall below, CLOMPR involves several modifications to OMPR.

\begin{itemize}

\item \noindent{\bf Non-negativity.} The compressive mixture estimation framework imposes a non-negativity constraint on the weights $\valpha$. Thus {\bf step 1} maximizes the real part of the correlation instead of its modulus. Similarly, in {\bf step 4} a Non-Negative Least-Squares minimization is performed instead of a classical Least-Squares minimization.

\item \noindent{\bf Continuous dictionary.} The dictionary $\{\skop\dirac_{\vc}\}$ is continuously indexed and cannot be exhaustively searched. The maximization in {\bf step 1} is thus done with a gradient ascent $\texttt{maximize}_{\vc}$, leading to a --local-- maximum of the correlation between atom and residual.

\item \noindent Compared to OMP, CLOMPR involves an additional gradient descent $\texttt{minimize}_{\support,\valpha}$ \emph{initialized with the current parameters} ({\bf step 5}), to further reduce the cost function \eqref{eq:pbm_skop} (but also leading to a \emph{local} minimum of \eqref{eq:pbm_skop}).
\end{itemize}
We also bring some modifications to the original CLOMPR \cite{Keriven2015,Keriven2016}:
\begin{itemize}
\item \noindent {\bf Initialization strategies.} We test several initialization strategies for {\bf step 1}, each being somehow similar to usual initialization strategies for $K$-means, see Section \ref{sec:init}.
\item \noindent {\bf Additional constraints.} We add constraints to the gradient descents. During the computation of the sketch $\hat\vz$ we also compute bounds $\vl,~\vu \in \Rbb^n$ such that all data are comprised in these bounds: denoting $\vec{a}\leq\vec{b}$ the element-by-element comparison of vectors in $\Rbb^n$, $\vl$ and $\vu$ are such that $\vl \leq \vx_i \leq \vu$ for all $i$'s. Note that the computation of these bounds is also done in one pass over $\dataset$. Then we enforce $\vl \leq \vc\leq \vu$ in $\texttt{maximize}_{\vc}$ and $\vl \leq \vc_k\leq \vu$ for all $k$'s in $\texttt{minimize}_{\support,\valpha}$.

\end{itemize}

\noindent We denote the resulting algorithm Compressive $K$-means (CKM).



\begingroup
\removelatexerror
\begin{algorithm*}[H]
\SetKwBlock{uBegin}{Step}{end}
\KwData{Sketch $\hat\vz$, frequencies $\freqs$, parameter $K$ , bounds $\vl,~\vu$}
\KwResult{Centroids $\support$, weights $\valpha$}
$\hat\vr \leftarrow \hat\vz$; $\support \leftarrow \emptyset$ \;
\For{$t \leftarrow 1$ \KwTo $2K$}{
\uBegin({\bf 1}: Find a new centroid){\nosemic$\vc \leftarrow \texttt{maximize}_{\vc} \left(\mathrm{Re}\scp{\frac{\skop \dirac_{\vc}}{\norm{\skop \dirac_{\vc}}}}{\hat\vr},~\vl,~\vu\right)$ \;
}
\uBegin({\bf 2}: Expand support){\nosemic $\support \leftarrow \support\cup \{\vc\}$ \;
}
\uBegin({\bf 3}: Enforce sparsity by Hard Thresholding if $t>K$){\If{$|\support|>K$}{\nosemic
$\boldsymbol\beta \leftarrow \arg\min_{\boldsymbol\beta \geq 0} \norm{\hat\vz - \sum_{k=1}^{|\support|} \beta_k \frac{\skop \dirac_{\vc_k}}{\norm{\skop \dirac_{\vc_k}}}}$\;
\nosemic Select $K$ largest entries $\beta_{i_1},...,\beta_{i_K}$\;
\nosemic Reduce the support $\support \leftarrow \{\vc_{i_1},...,\vc_{i_K}\}$\;
}
}
\uBegin({\bf 4}: Project to find $\valpha$){\nosemic $\valpha \leftarrow \arg\min_{\valpha \geq 0} \norm{\hat\vz - \sum_{k=1}^{|\support|} \alpha_k \skop \dirac_{\vc_k}}$\;
}
\uBegin({\bf 5}: Global gradient descent){\nosemic $\support,\valpha \leftarrow \texttt{minimize}_{\support,\valpha}\Big( \norm{\hat\vz - \sum_{k=1}^{|\support|} \alpha_k \skop \dirac_{\vc_k}},~\vl,~\vu\Big)$\;
}
\nosemic Update residual: $\hat\vr \leftarrow \hat\vz - \sum_{k=1}^{|\support|} {\alpha}_k \skop \dirac_{\vc_k}$\;
}
\caption{CLOMPR for $K$-means (CKM)}
 \label{algo:OMPR}
\end{algorithm*}
\endgroup

%
%
%
\subsection{Complexity of the method}
%
%

The method can be summarized as follows. Given a dataset $\dataset$, a number of frequencies $m$ and a number of clusters $K$,
\begin{enumerate}
\item Use the algorithm in \cite{Keriven2016} on a small fraction of $\dataset$ to choose a frequency distribution $\freqdist$;
\item Draw $m$ frequencies $\freq_j$ $i.i.d.$ from $\freqdist$;
\item compute the sketch $\skop \hat p = \Big[\frac{1}{N}\sum_{i=1}^N e^{-i\freq_j^T\vx_i}\Big]_{j=1}^m$;
\item Retrieve $\support$ from the sketch using the CKM algorithm.
\end{enumerate}

The CKM algorithm scales in $\order{K^2mn}$, which is potentially far lower than the $\order{nNKI}$ of Lloyd-Max 
for large $N$.

To compute the sketch, one has to perform the multiplication $\mW^T\mX$, where $\mX=[\vx_1,...,\vx_N]$ and $\mW=[\freq_1,...,\freq_m]$ are the matrices of data and frequencies. It theoretically scales in $\order{nmN}$, but can be done in a distributed manner by splitting the dataset over several computing units and averaging the obtained sketches, such that the full data need never be stored in one single location. One can also exploit GPU computing for very large-scale matrix multiplication \cite{Zhang2015}. The proposed sketch can also be maintained online, which is another crucial property of typical dataset sketches \cite{Cormode2011}.

Some techniques might further reduce these complexities. As detailed in~\cite{Keriven2016toolbox}, most operations in CKM can be narrowed down to performing multiplications by $\mW$ and $\mW^T$. Therefore, both computing the sketch and performing CKM could benefit from the replacement of $\mW$ by a suitably randomized fast transform
\cite{Do2011,Lemagoarou2015}. 

It is also possible to reduce the dimension $n$ to $\order{\log K}$ with random projections \cite{Boutsidis2010}, as a preprocessing step.

Finally, empirical results (see Sec.~\ref{sec:pt}) suggest that 
the size of the sketch only needs to scale linearly with the number of parameters, i.e. $m\approx \order{nK}$. 
Combining all these results, it would be potentially possible to compute the sketch in $\order{KNT^{-1}(\log K)^2}$, where $T$ is the number of parallel computing units, and perform CKM in $\order{K^3(\log K)^2}$. We also mention that a hierarchical adaptation of CLOMPR which scales in $\order{K^2(\log K)^3}$ has been proposed for GMM estimation \cite{Keriven2016}, and that a variant for the $K$-means setting considered here might be implementable. We leave possible integration of those techniques to future work.

\section{Experiments}\label{sec:experiments}

\subsection{Setup}

We compare our Matlab implementation of CKM, available at~\cite{Keriven2016toolbox}, with Matlab's \texttt{kmeans} function that implements Lloyd-Max.

We first use artificial clustered data drawn from a mixture of $K$ unit Gaussians in 
dimension $n$ with uniform weights, with means $\vec{\mu_k}$ drawn according to a centered 
Gaussian with covariance $cK^{1/n}\ma{Id}$. The constant $c=1.5$ is chosen so that clusters are sufficiently 
separated with high probability. Unless indicated otherwise, $N=3\cdot 10^5$ points are generated 
from $K=10$ clusters with $n=10$.

The second problem consists in performing spectral clustering~\cite{Uw2001}  on the MNIST 
dataset~\cite{Lecun1998}. In fact, to test our method's performance on a large dataset, 
we use the original $7\cdot 10^4$ images, that we complete with images artificially 
created by distortion of the original ones using the toolbox infMNIST proposed in~\cite{Loosli2007}. 
We thereby test on three dataset sizes: 
the original one with $N_1=7\cdot 10^4$ and two augmented ones with $N_2=3\cdot 10^5,~N_3=10^6$. 
For each dataset, 
we extract SIFT~\cite{Vedaldi2010} descriptors of each image, and compute the $K$-nearest neighbours adjacency matrix 
 (with $K=10$) using FLANN~\cite{Muja2009}. As we know there are ten classes, we compute the 
first ten eigenvectors of the associated normalized Laplacian matrix, and run CKM on these $N$ 
10-dimensional feature vectors. 
Note that spectral clustering 
requires the first few eigenvectors of the global Laplacian matrix, of size $N^{2}$, which becomes prohibitive for large $N$. Replacing the $K$-means step by CKM 
in compressive versions of spectral clustering~\cite{Tremblay2016, Tremblay2016a} or in efficient 
kernel methods such as in~\cite{Chitta2012} are left for future 
investigations.

Unless indicated otherwise $m=1000$ frequencies are used. Each result is averaged over 100 experiments.

\subsection{Initialization strategies} 
\label{sec:init}
Several strategies to initialize the gradient descent $\texttt{maximize}_{\vc}$ in {\bf step 1} of CKM are tested, along with their equivalents in the usual \texttt{kmeans} 
algorithm. Note that, for \texttt{kmeans} 
it corresponds to selecting $K$ initial centroids then running the algorithm, while for CKM each iteration is initialized with a single new centroid.
\begin{itemize}


\item \noindent {\bf Range:} for CKM, pick $\vc$ where each component $c_i$ is drawn uniformly at random with $l_i \leq c_i \leq u_i$; for \texttt{kmeans},
select $K$ such points.

\item \noindent {\bf Sample:} for CKM, select a point $\vc=\vx_i$ from the data at random; for \texttt{kmeans}, 
select $K$ such points.

\item \noindent {\bf K++,}  a strategy analog to the $K$-means++ algorithm \cite{Arthur2007}: for CKM, select $\vc=\vx_i$ from the data with a probability inversely proportional to its distance to the \emph{current} set of centroids $\support$; for \texttt{kmeans}, 
run exactly the K++ algorithm \cite{Arthur2007}.
\end{itemize}
NB: the last two strategies do not exactly fit in the "compressive" framework, where data are sketched then discarded, since they still require access to the data. They are implemented for testing purpose.

\begin{figure}
\begin{center}
\includegraphics[scale=\sca]{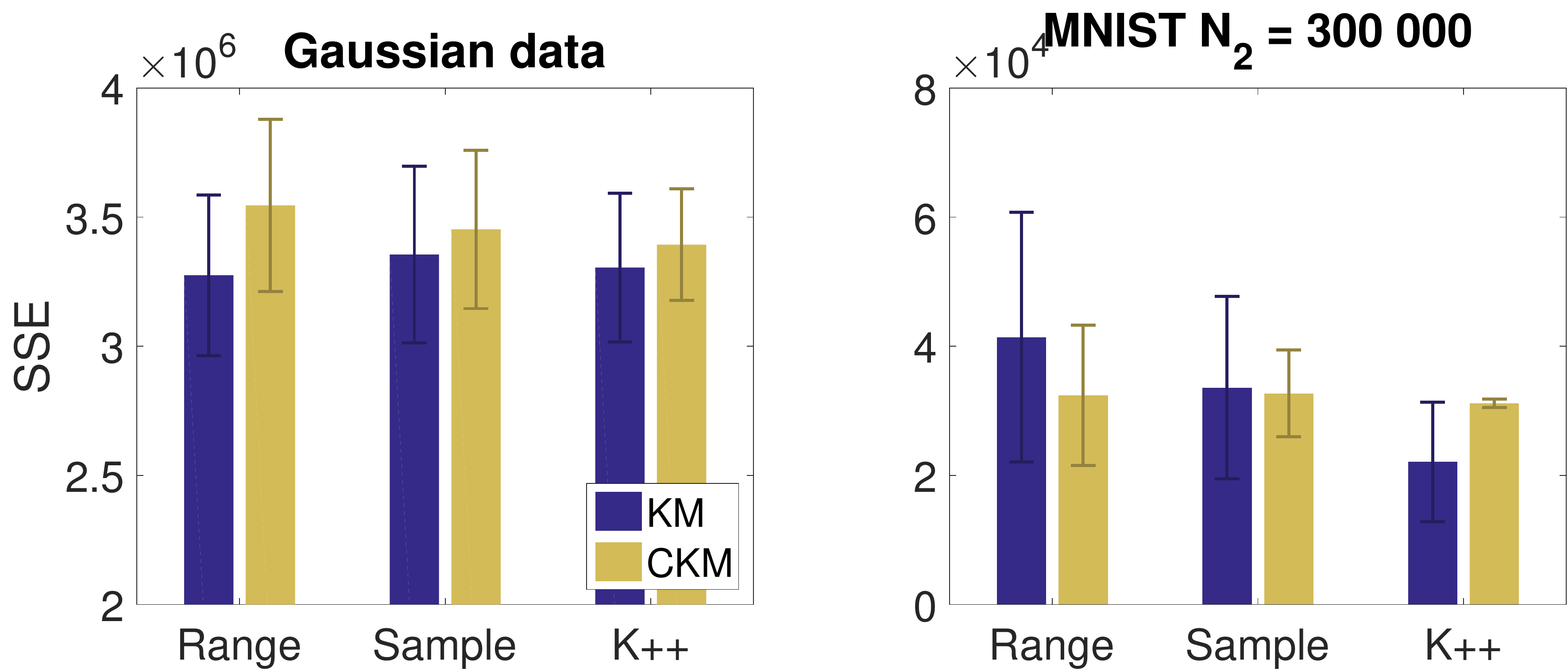}
\end{center}
\vspace{\vsp}
\caption{Comparison of initialization strategies. Mean and variance of SSE cost over 100 experiments.}
\vspace{\vsp}
\label{fig:init}
\end{figure}

In Fig.~\ref{fig:init} the different initialization strategies are compared, by displaying the mean and variance of the SSE over 100 experiments. On Gaussian data, both algorithms yield approximately the same SSE for all strategies. On MNIST data, for CKM all strategies approximately yield the same result, but the Sample and K++ initializations give a lower variance. The \texttt{kmeans} 
 algorithm is more sensitive to initialization, and only outperforms the CKM algorithm for the K++ strategy. In all further experiments the ``Range" strategy is used for both algorithms.

\subsection{Number of frequencies}
\label{sec:pt}

\begin{figure}
\begin{center}
\includegraphics[scale=\sca]{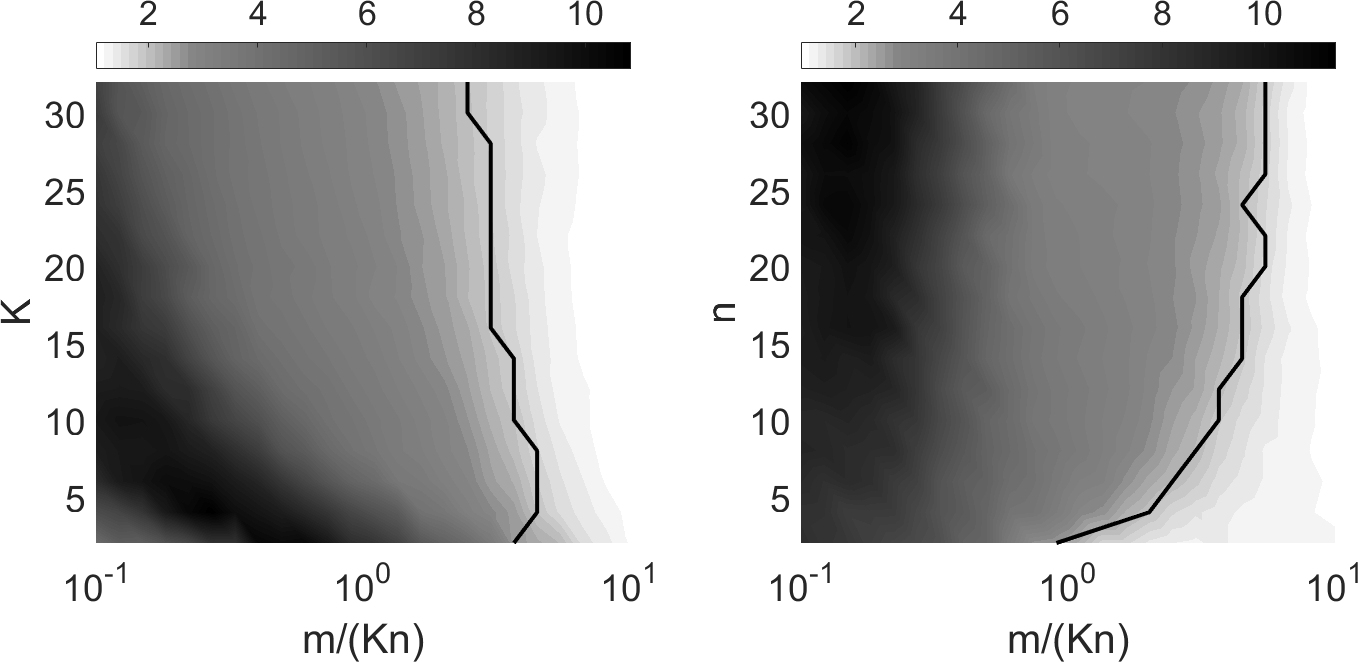}
\end{center}
\vspace{\vsp}
\caption{Relative SSE ($i.e.$ SSE obtained with CKM divided by that obtained with \texttt{kmeans}
) on Gaussian data, with $n=10$ (left) and $K=10$ (right). Lines are drawn where the relative SSE becomes lower than $2$.}
\label{fig:pt}
\end{figure}

It is interesting to empirically evaluate how many frequencies $m$ are required for CKM to be effective. 
In Fig~\ref{fig:pt}, 
we show the SSE obtained with CKM divided by that of \texttt{kmeans} 
with respect to the relative number 
of frequencies $m/(Kn)$, and draw lines where the relative SSE becomes lower than $2$. It is seen that 
these lines are almost constant at $m/(Kn) \approx 5$, except for a deviation at low $n$. Recent, 
preliminary theoretical results on GMMs \cite{Keriven2016} hint that for a fixed error level 
the required number of frequencies grows proportional to the number of parameters $m\approx \order{Kn}$. We postulate that the same phenomenon might be valid for $K$-means clustering.

\subsection{Scalability and performance of CKM}

\begin{figure}[t]
\begin{center}
\includegraphics[scale=\sca]{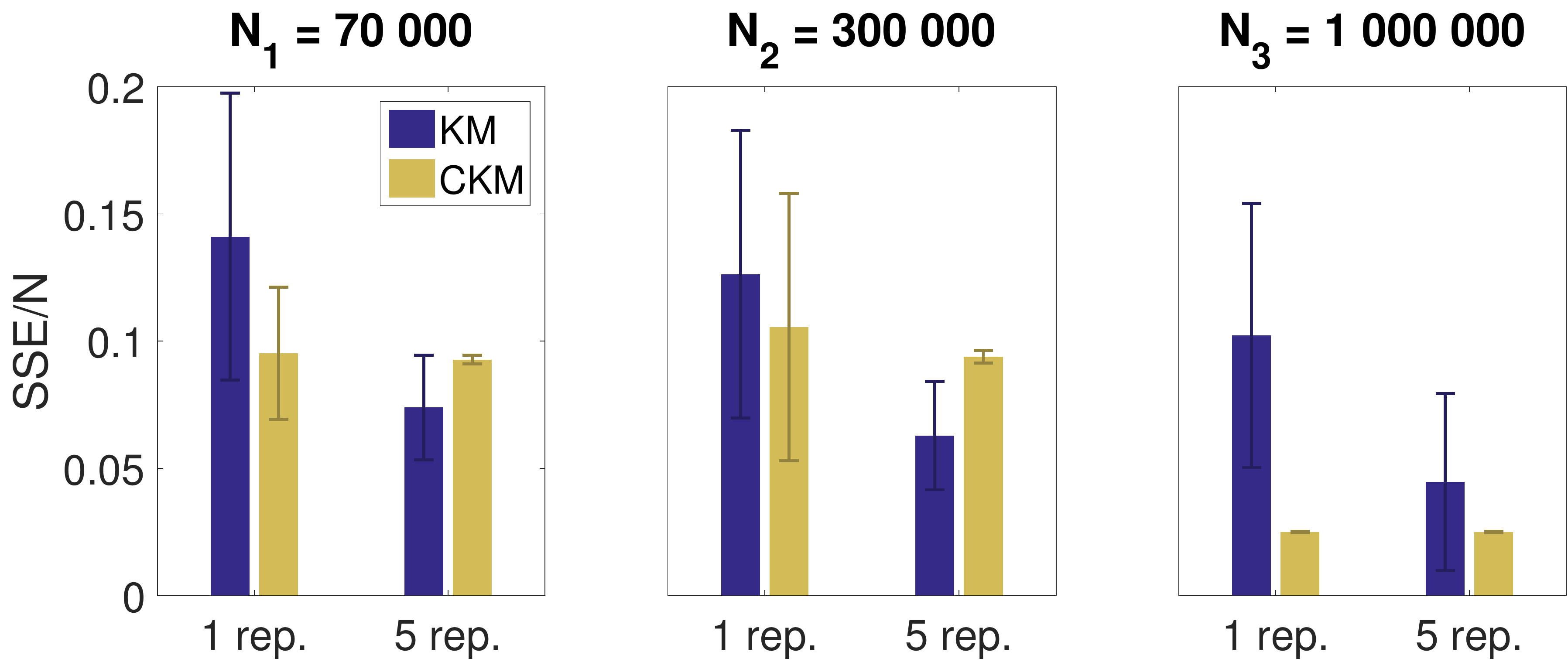} \\
\includegraphics[scale=\sca]{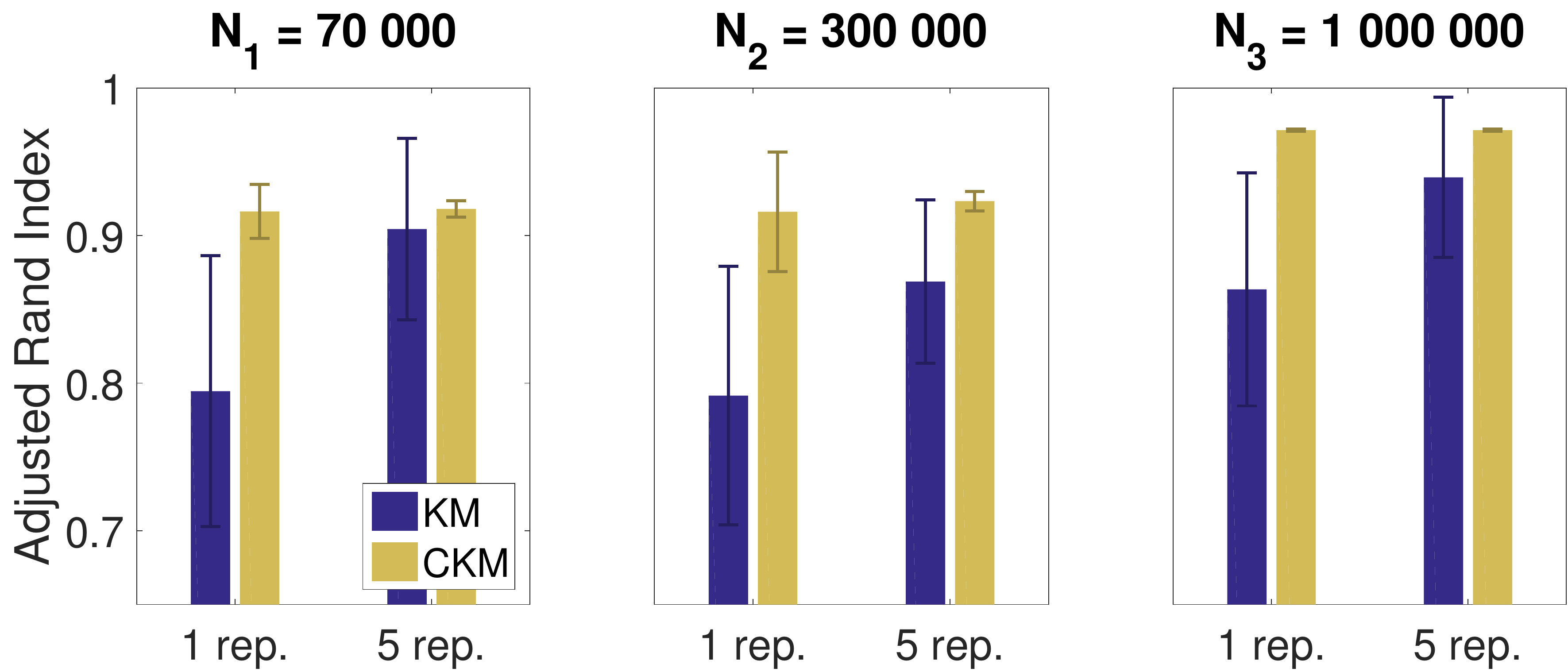}
\end{center}
\vspace{\vsp}
\caption{Mean and variance over 100 experiments of SSE divided by $N$ (top, lower is better) and Adjusted Rand Index \cite{Rand1971} for comparing clustering results (bottom, higher is better) on MNIST, for $1$ or $5$ replicates.}
\vspace{\vsp}
\label{fig:mnist}
\end{figure}


Often Lloyd-Max 
is repeated several times with random initializations, and the set of centroids yielding the lowest SSE is kept. In the CKM algorithm, we do not have access to the SSE in practice since the data are discarded after computation of the sketch. Hence, when several replicates of CKM are performed, \emph{we select instead the set of centroids that minimizes the cost function} \eqref{eq:pbm_skop}. 


In Fig.~\ref{fig:mnist}, we evaluate the SSE and classification performance on the MNIST dataset, for $1$ or $5$ replicates of the algorithms. As expected, \texttt{kmeans}
greatly benefits from being performed several times, while CKM is more stable between number of replicates. This allows CKM to be run with (much) fewer replicates than \texttt{kmeans} 
in practice.
Moreover, for a large dataset ($N_3=10^{6}$), the performance of CKM has negligible variance and negligible difference between $1$ and $5$ replicates. 
Hence, \emph{although the size $m$ of the sketch is kept fixed for all $N$'s, the method is actually more efficient when applied to large datasets}.

Interestingly, in all cases CKM outperforms \texttt{kmeans} 
in terms of classification (Fig. \ref{fig:mnist}, bottom). This might mean that the proposed cost function is more adapted than the SSE on this particular task.

\begin{figure}[t]
\begin{center}
\includegraphics[scale=\sca]{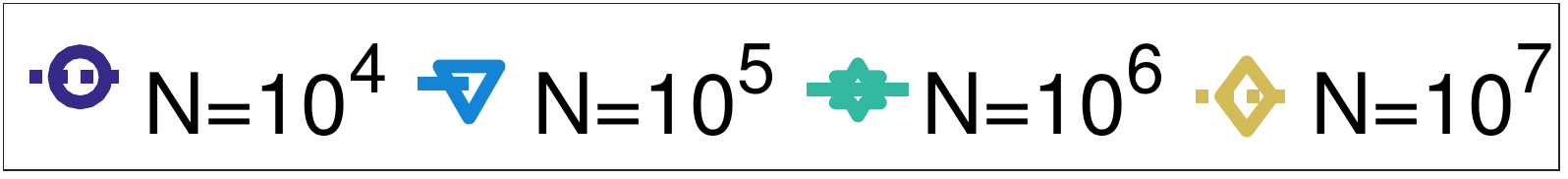} \\
\includegraphics[scale=\sca]{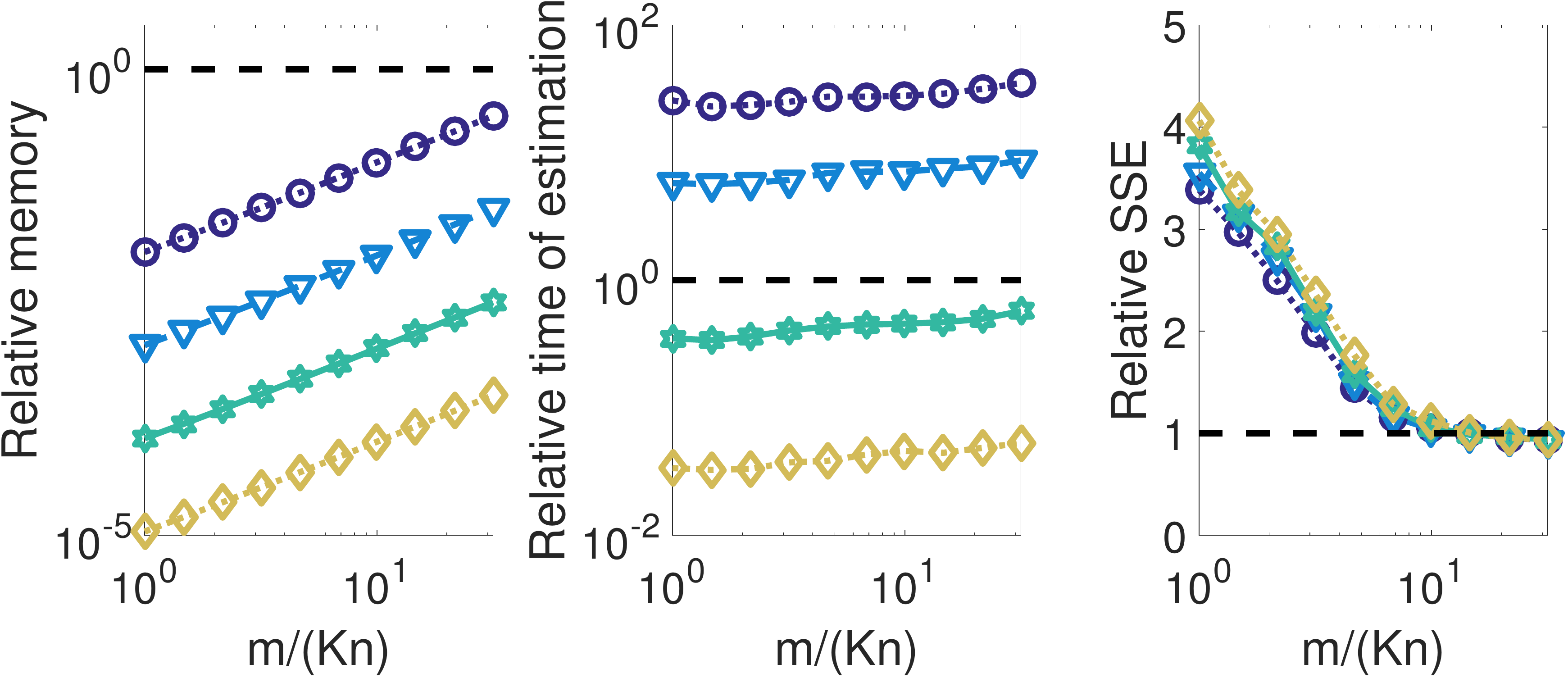}
\end{center}
\vspace{\vsp}
\caption{Relative time, memory and SSE of CKM algorithm with respect to \emph{one run} of \texttt{kmeans} 
($10^0$ represent the \texttt{kmeans} 
result), on Gaussian data.}
\vspace{\vsp}
\label{fig:time}
\end{figure}

We finally examine the time and memory complexities of CKM, relatively to that of \texttt{kmeans}, 
in Fig.~\ref{fig:time}. Note that the computational complexity of computing the sketch is not outlined on this figure, since it can be done in an online and massively parallelized manner and is highly dependent on the user's available hardware. As expected, given the sketch CKM is far more efficient than \texttt{kmeans} 
on large datasets, even for a high number of frequencies.
Overall, on a dataset with $10^7$ elements, one run of CKM is up to $150$ times faster than \texttt{kmeans} 
with $5$ replicates.

%
%

\section{Conclusion}

We presented a method for performing $K$-means on large datasets, where the centroids are derived from a \emph{sketch} of the data. The problem was linked to generalized Compressive Sensing, where a probability distribution is measured through a linear operator and reconstructed as a sparse mixture of $K$ Diracs. A modified version of the CLOMPR algorithm \cite{Keriven2015, Keriven2016} is used for estimating this mixture.

Results showed that, although the proposed objective is not directly linked to the traditional SSE cost, our method compares favorably with usual algorithms for $K$-means. It is much more stable to initialization, and generally succeeds with only one replicate. Although the size of the sketch does not depend on $N$, compared to usual $K$-means the proposed algorithm is all the more effective when applied on large datasets, in terms of complexity, SSE and classification performance on the MNIST dataset for instance.

\noindent {\bf Outlooks} As already mentioned, it is possible to combine the proposed approach with dimension reduction \cite{Boutsidis2010} and/or fast transforms \cite{Lemagoarou2015} to speed-up the method even more.

The proposed method was observed to outperform usual $K$-means for classification on the MNIST dataset, even when it performs worse in terms of SSE. These encouraging results may lead to further engineering of objective functions and innovative clustering methods. As mentioned in the introduction, the cost function \eqref{eq:pbm_skop} can be connected to a finite embedding of probability distributions in Hilbert Spaces \cite{Sriperumbudur2010} with Random Features \cite{Rahimi2007}. In this framework, theoretical results about the information preservation property of sketches have been derived on GMMs \cite{Keriven2016}, and further work will examine such results in the context of mixtures of Diracs.

\bibliographystyle{IEEEbib}
\clearpage
\bibliography{biblio_cleaned}

\end{document}

%% file: macro+variables.tex
\newcommand{\mX}{\ensuremath{\ma{X}}}
\newcommand{\mW}{\ensuremath{\ma{W}}}
\newcommand{\sk}{\ensuremath{\ma{Sk}}}

\newcommand{\vone}{\ensuremath{\vec{1}}}
\newcommand{\vx}{\ensuremath{\vec{x}}}
\newcommand{\vy}{\ensuremath{\vec{y}}}
\newcommand{\vz}{\ensuremath{\vec{z}}}
\newcommand{\vr}{\ensuremath{\vec{r}}}
\newcommand{\vl}{\ensuremath{\vec{l}}}
\newcommand{\vu}{\ensuremath{\vec{u}}}
\newcommand{\valpha}{\ensuremath{\boldsymbol\alpha}}
\newcommand{\vc}{\ensuremath{\vec{c}}}
\newcommand{\support}{C}

\newcommand{\dataset}{X}
\newcommand{\freq}{\ensuremath{\boldsymbol\omega}}
\newcommand{\freqs}{\ensuremath{\Omega}}
\newcommand{\freqdist}{\ensuremath{\Lambda}}
\newcommand{\dirac}{\ensuremath{\boldsymbol \delta}}

\newcommand{\skop}{\ensuremath{\mathcal{A}}}

\newcommand{\Cbb}{\ensuremath{\mathbb{C}}} 
\newcommand{\Rbb}{\ensuremath{\mathbb{R}}} 

\newcommand{\Ebb}{\ensuremath{\mathbb{E}}}

\newcommand{\scp}[2]{\ensuremath{\left\langle #1, #2 \right\rangle}}

\newcommand{\norm}[1]{\ensuremath{\left\| #1\right\|}}

\newcommand{\ma}[1]{\ensuremath{\mathsf{#1}}}
\renewcommand{\vec}[1]{\ensuremath{\bm{#1}}}

\DeclareMathOperator*{\argmin}{argmin}

\newcommand{\order}[1]{\ensuremath{\mathcal{O}\left(#1\right)}}
